# An Enhanced Hierarchical Planning Framework for Multi-Robot Autonomous Exploration

Gengyuan Cai*, Xiangmao Chang, Luosong Guo

*College of Computer Science and Techonology
Nanjing University of Aeronautics and Astronautics, Nanjing 210016, China
Email: gengyuanc@nuaa.edu.cn, xiangmaoch@nuaa.edu.cn, luosongguo@nuaa.edu.cn

*Abstract*

**The autonomous exploration of environments by multi-robot systems is a critical task with broad applications in rescue missions, exploration endeavors, and beyond. Current approaches often rely on either greedy frontier selection or end-to-end deep reinforcement learning (DRL) methods, yet these methods are frequently hampered by limitations such as short-sightedness, overlooking long-term implications, and convergence difficulties stemming from the intricate high-dimensional learning space. To address these challenges, this paper introduces an innovative integration strategy that combines the low-dimensional action space efficiency of frontier-based methods with the far-sightedness and optimality of DRL-based approaches. We propose a three-tiered planning framework that first identifies frontiers in free space, creating a sparse map representation that lightens data transmission burdens and reduces the DRL action space's dimensionality. Subsequently, we develop a multi-graph neural network (mGNN) that incorporates states of potential targets and robots, leveraging policy-based reinforcement learning to compute affinities, thereby superseding traditional heuristic utility values. Lastly, we implement local routing planning through subsequence search, which avoids exhaustive sequence traversal. Extensive validation across diverse scenarios and comprehensive simulation results demonstrate the effectiveness of our proposed method. Compared to baseline approaches, our framework achieves environmental exploration with fewer time steps and a notable reduction of over 30% in data transmission, showcasing its superiority in terms of efficiency and performance.**

## I. INTRODUCTION

Exploring unknown environments is a fundamental task in robotics applications. It involves observing and mapping the environment using sensor data without prior knowledge of the surroundings. Simultaneously, it estimates its own state and makes decisions and plans for the exploration task accordingly [1]. The goal is to scan the specified unknown environment as thoroughly and accurately as possible in the shortest time, to establish conditions for subsequent robot tasks [7]. This has a wide range of applications, including search and rescue, exploration, and cleaning robots [2].

Significant work has been produced in the field of robot exploration tasks. The most well-known pioneering work is based on frontier points [8]. This method identifies frontier points between free spaces and unknown spaces in the map as target points for exploration tasks, uses Shannon entropy to evaluate environmental uncertainty, considering the cost in the planning process to select the best point for the next step [4]. This method is simple, easy to implement, and computationally efficient for small clusters of mobile robots. This method relies on specific features and has limited adaptability, and most implementations use heuristic or meta-heuristic optimizations, which can lead to short-sightedness and suboptimal decision-making [3]. Especially for large clusters of robots, the computational requirements increase exponentially.

Reinforcement learning (RL)-based approaches have gained increasing attention in recent years for their ability to discover near-optimal solutions for exploring unknown spaces [9]-[10]. In this context, the exploration task involves selecting a series of actions based on noisy sensor measurements in a partially observable environment, aiming to reduce uncertainty in state estimation and mapping. This can be modeled as a partially observable Markov decision process (POMDP). To address the dimensionality problem in traditional reinforcement learning, these approaches typically use environmental information as the state and directly output control actions through end-to-end neural networks to drive the robot's exploration [13]-[14]. This approach is widely studied and used due to its optimality and adaptability. However, this method still has certain limitations. Searching the entire action space requires a long time to converge, especially in multi-robot systems. Due to its end-to-end control decisions, the robot's exploration decision and planning are strongly coupled, resulting in low adaptability when switching between different control-driven unmanned robots or migrating the algorithm from simulation to the real world. In order to address these issues, some methods establish the environmental topology and plan exploration tasks using graph neural networks (GCN). However, topology-based exploration tends to ignore some details, leading to incomplete mapping in complex environments. Or using a local graph instead of a global graph for planning can accelerate convergence while may lead to suboptimal planning outcomes.

With advancements in sensors, robotics, and Simultaneous Localization and Mapping (SLAM), this paradigm has been gradually extended to multi-robot systems [5]. Compared to single-robot systems, exploration using multi-robot systems (MRS) offers higher efficiency and robustness. However, as the number of robots increases, task allocation and planning require more complex decision-making algorithms. Deep reinforcement learning (DRL)-based methods also face the challenge of exponential growth in spatial dimensions. In centralized multi-robot systems, the computing center needs to receive environmental data from each robot to calculate the action space for DRL. Most existing multi-robot exploration systems integrate local information by acquiring sensor data or complete map data, but this consumes significant communication resources, especially in systems using 3D LiDAR [6]. Thus, enabling the computing center to obtain the required data with lower transmission bandwidth is an urgent problem.

In order to solve these problems, this paper proposes a hierarchical planning method combining frontier-based method and reinforcement learning-based method. The motivation is to reduce the action space of DRL by frontier-based method, and use multi-Graph Neural Network (mGNN)-based Proximal Policy Optimization (PPO) to improve the optimality and generalization of exploration area allocation, and comprehensively consider the exploration objectives in the region to obtain more reasonable local planning routing. Specifically, the main contributions of this paper are as follows:
1) We proposed a three-layer exploration planning architecture, including the preprocessing layer, the global planning layer, and the local planning layer, which combines the advantages of frontiers-based methods and learning-based methods, addressing the optimization problems and the dimensionality issues of the action space in traditional approaches. Additionally, we designed a lightweight data structure to reduce network transmission pressure in multi-robot systems.
2) The mGNN and PPO strategy are used to calculate the affinity of the target area as the target utility for allocation. The utility function is used locally and the optimal route is calculated by subsequence inversion to obtain the target point and control.
3) Through a large number of simulation experiments, it is verified that our method can consume fewer time steps than the baseline method while achieving the same goal in the small, middle and large scare scenes. And thanks to our designed data structure, data transmission volume has been reduced by more than 30%.

## II. RELATED WORK

### A. Multi-robot automatic exploration

Most previous work on autonomous exploration has focused on single-robot scenarios. As a pioneering work in the field of active robotic exploration, [7] introduced the concept of frontiers, which are the frontiers between free space and unknown areas, and used these frontier points as targets for each step of the exploration task. [8] extended this idea to 3D scenes, using unknown voxel frontiers as target points. Autonomous exploration also relies on information theory, selecting frontiers based on immediate information gain.

Unlike the traditional methods mentioned above that primarily use short-sighted strategies to determine target locations from a set of frontiers, recent approaches [9-10] use the state of the map as the main input and employ convolutional neural networks as decoders to compute actions with long-term value through reinforcement learning. Other work explores the environment by constructing topological maps [11] or semantic maps, while these methods may overlook certain details, leading to incomplete mapping.

For multi-robot systems, many methods are similar to those used in single-robot systems, with the primary difference being how to assign targets to multiple robots to maximize global long-term benefits. An intuitive approach is to rank candidate targets by their geodesic distance as utility and then assign them to robots in a sequential manner. Current methods [16] mainly rely on solving optimal mass transportation problems between robots and targets, such as the multi-traveling salesman problem for target allocation. According to the analysis in [12], the aforementioned methods select target locations from the nearest targets under multi-robot coordination constraints, exhibiting short-sightedness. This paper proposes learning better neural distances through reinforcement learning to achieve more efficient map construction.

### B. Deep reinforcement learning

To address the optimality issue in discrete spaces and the dimensionality problem in continuous spaces within reinforcement learning, numerous studies have introduced neural networks to map between continuous and discrete spaces. [13] proposed a potential dynamic Q-learning method that combines Q-learning with artificial potential fields and dynamic reward functions to generate feasible paths with shorter lengths and smaller turning angles. [14] proposes a solution for Active Simultaneous Localization and Mapping (Active SLAM) of robots in unknown indoor environments using a combination of Deep Deterministic Policy Gradient (DDPG) path planning and the Cartographer algorithm.

Despite the aforementioned efforts to enhance optimality and avoid dimensionality disasters in path planning within continuous spaces, most RL-based methods perform optimal decision search across the entire action space, resulting in a prolonged training process to achieve convergence. Even though some RL-based methods reduce the search space to a limited set of specific local maps, the selection of certain values remains suboptimal, leading to numerous futile trial-and-error attempts during the training phase. Furthermore, while using partial local maps can accelerate training compared to utilizing the entire map, it may cause the robot to become trapped in local optima due to shortsightedness. Consequently, [15] proposed a hybrid approach combining frontier point-based methods with learning-based methods to reduce the action space of reinforcement learning methods from the entire map to a few points of interest, addressing the lengthy training time of existing reinforcement learning methods and the suboptimal issues in traditional frontier-based methods. However, this approach is still limited to single-robot systems. [17] simplified the multi-robot active mapping problem into bipartite graph matching, establishing node correspondences between two graphs representing robots and frontiers, respectively, and used multi-graph neural networks to learn the affinity matrix to select the optimal target points. Unlike the aforementioned work, our research will focus on multi-robot systems, integrating frontier-based methods with learning-based approaches, and designing optimal routing in the local planning layer to achieve a reasonable plan for both long-term and short-term goals.

## III. SYSTEM MODEL AND PROBLEM FORMULATION

### A. System Model

**Environmental Model:** Environmental Model: Our environment is constructed by acquiring point clouds through LiDAR perception. To facilitate map exploration calculations, the point cloud map is rasterized into an octomap. This involves recursively dividing a cube into 8 child nodes until a leaf node is reached. The probability expresses whether a leaf is occupied by a point cloud. If the threshold is met, it is considered occupied. Formally, for $t = 1, 2, ..., T$, the observed data is $z_1, z_2, ..., z_T$. The probability of the :

$$P(n|z_{1:T}) = \left[1 + \frac{1-P(n|z_T)}{P(n|z_T)} \cdot \frac{1-P(n|z_{1:T-1})}{P(n|z_{1:T-1})} \cdot \frac{P(n)}{1-P(n)}\right], \quad (1)$$

where $P(n|z_{1:T})$ represents the occupancy probability of the $n_{th}$ child node calculated based on the observed data from time 1 to $T$. After obtaining the octree map, by projecting the voxels within the specified range onto the $xy$ plane, the grid in the three-dimensional space can be converted into a two-dimensional grid map $G$ with a height of $h$, a width of $w$, and a grid resolution of $r$ by projection, which is divided into free space, unknown space, and occupied space according to the occupancy situation. With this we can define the frontier point model, for a free grid $g_{ij}$ in the grid map $G$, if there are unknown grids around it, i.e., $g_{i-1,j}, g_{i+1,j}, g_{i,j-1}, g_{i,j+1}$, we have $g_{ij} \in F_{un} \subset \mathbb{R}^3$. Similarly, when there are occupied grids around it, we have $g_{ij} \in F_{occ} \subset \mathbb{R}^3$, which includes the grid coordinates $x, y$, and the frontier label. For robot $r_i \in R (1 \leq i \leq N)$, with $s_{ri} \in \mathbb{R}^3$ to represent the state of the robot, which includes the $x, y$ coordinates and the robot label relative to the origin of the map.

**Multi Graph Neural Network:** Based on the reconstructed grid map and the cluster center positions of the robot and frontier points, we construct the robot Graph $G_r = (V_r, E_r)$ and the candidate cluster center Graph $G_c = (V_c, E_c)$, as well as a cross graph $G_{rc} = (V_r, V_c, E_{rc})$ connecting the robot and the cluster center. Where $V_r$ and $E_r$ represent the features of the robots and the weights of the relationships between them, while $V_c$ and $E_c$ represent the features of the cluster centers and the weights of the relationships between them. $G_{rc}$ is the bipartite graph connecting all $V_c$ and $V_r$, $E_{rc}$ represents the matching utility between the robot and the cluster center to be learned. Later, we utilize a mGNN and PPO to calculate the edges of the graph, which represent the relationships between the vertices. The aim is to maximize the utility of the two allocations by computing the edge relationships within the bipartite graph:

$$E^* = max \sum_{r=1}^{n} E_{rc}. \quad (2)$$

**Local Planning Routing:** For the routing planning of local target points cluster $C_k$, define $\rho_k = [f_{k0}, f_{k1}, f_{k2}, ..., f_{kn}]$ as the sequence of all candidate points, where $f_{kj}$ represents a member of cluster $C_k$, $f_{k0}$ as the current position of the robot, and the optimal path $\rho_k^* = [f_{k0}, f^*]$. By considering all candidate points in the optimization, a more efficient routing is obtained. $U(f_{kj})$ is the utility of the candidate point $f_{kj}$, and the goal is to maximize the total utility:

$$U^* = max \sum_{j=1}^{n} U(f_{kj}). \quad (3)$$

### B. System Architecture

Our system architecture is shown in Fig. 1, which includes three layers of planning:

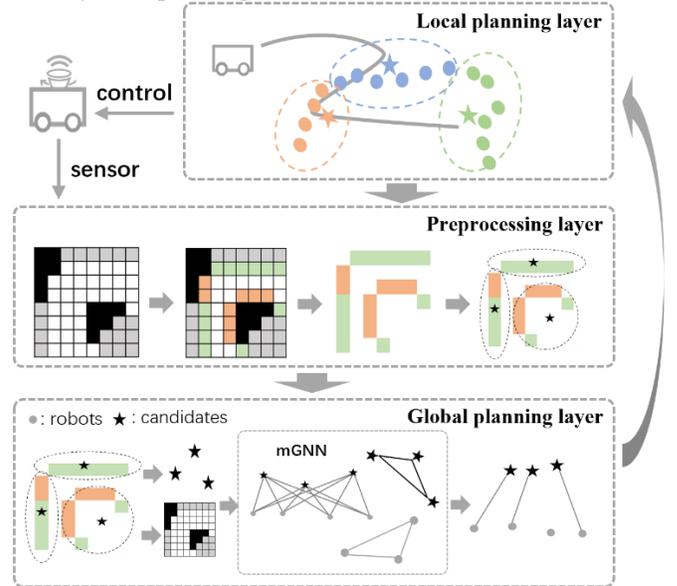

Fig. 1. System Architecture

**Preprocessing layer:** At this level, the global grid map and candidate target points are computed. Each robot $r_i$ obtains the local environment grid map and frontiers. Then at the computing center, we merge the local frontiers to obtain global frontiers $F_{un}$ and $F_{occ}$, and calculate the reconstructed global grid map $G$. The mean shift algorithm is used to obtain the cluster centers $c_k \in C_k \subset C$ and the cluster members $f_k \in C_k$. The cluster centers will be the candidate targets of the robots;

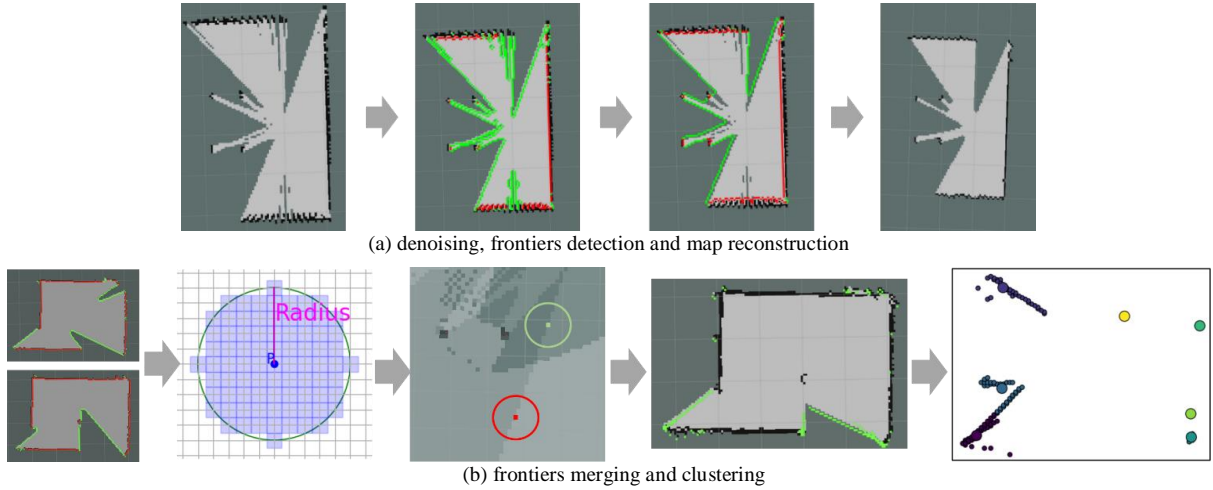

(a) denoising, frontiers detection and map reconstruction

(b) frontiers merging and clustering

Fig. 2. Environment Preprocessing Pipeline

**Global planning layer:** At this level, the assignment of robots to candidate points are calculated. According to the reconstructed global map $G$, the robot state $s_{ri}$ and the clustering center state $c_k$, the matching target position $c_k \sim s_{ri}$ is calculated by the cross graph;

**Local planning layer:** At this level, we perform local routing planning for the clustered members assigned to the robots. For $f_k \in C_k$, a routing $\rho^*$ that maximizes the utility $U$ is obtained, and then proceed with navigation and control to enable the robots to complete the task.

### C. Problem Formulation

Given an environment within a specified area, our goal is to complete its exploration in fewer time steps. To achieve this, each global goal selection must involve making decisions that comprehensively consider both cost and utility, thus facilitating subsequent exploration. Furthermore, in a multi-robot system, the exploration task should be accomplished with minimized data transmission to reduce the communication load on the system.

## IV. PROBLEM SOLUTION

We propose a novel integration of deep reinforcement learning and frontier-based methods, consisting of three stages: (i) preprocessing environmental information by extracting frontier points and identifying candidate target regions; (ii) calculating long-term utility and assigning candidate targets using a multi-graph neural network; and (iii) planning local routes using subsequence selection, algorithm 1 illustrates the solution proposed in this paper.

### A. Dense Frontiers Detection & Sparse Map Transmission

The grid map $G_i$ established by the SLAM algorithm often contains numerous noise points. Direct extraction of frontier points would result in high discreteness. Therefore, the free space is initially expanded using an dilation operation, and then the grid map is denoised using dilation and erosion operations, collectively referred to as a closing operation, as shown in Fig. 2.

For the denoised grid map $G_i$, a frontier extraction operation is performed. Traditional frontier extraction algorithms, such as the RRTs [19], can only obtain frontier points between free and unknown spaces and require significant computation for large areas. We propose extracting $F_{un}$ and $F_{occ}$ using multi-channel kernel convolution. First, $G_i$ is divided into three channel matrices $G_{\_un}$, $G_{\_occ}$, and $G_{\_free}$ using boolean indexing. Frontier points of the occupied and unknown spaces are then calculated by kernel convolution, followed by finding their intersection within the free space, as shown in Fig. 3. The extracted frontier points are dense, forming a closed area inside that supports subsequent map reconstruction at the center.

After completing the frontier detection, the local occupancy grid map can be converted into a sparse frontier map through the sparse graph structure and transmitted to the computing center. After obtaining the $F_i (1 \leq i \leq N)$ of each robot in the computing center, the sparse graph will be reconstructed into an occupancy grid map through the occupancy grid map reconstruction algorithm. Specifically, using the coordinates of $s_{ri}$ as the seed, the feasible area inside $F_{un\_i}$ and $F_{occ\_i}$ is filled with free space using the scan-line filling algorithm. Since the position of the current robot $s_{ri}$ is used as the seed, some external frontier noise points can be in unreachable areas. After removing the noise points using the closing operation, the grid map constructed by the sparse graph $G'$ can be obtained.

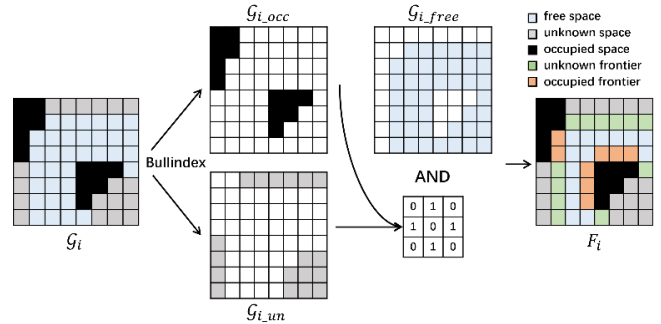

Fig. 3. Three-channel Occupancy Grid Map Frontiers Detection

**Algorithm 1** Multi-Robot Collaborative Exploration

**Input:** $G, s_{ri}, N$
**Output:** target point and control
1: **for** $r_i \in R$ **do**:
2:     $F_{un\_i}, F_{occ\_i} = frontiers\_detection(G, s_{ri})$
3:     transmit $F_{un\_i}, F_{occ\_i}, s_{ri}$ to center
4:     $F_{un}, F_{occ} =$ merge frontiers $F_{un\_i}$ and $F_{occ\_i}$
5: **end for**
6: $G' =$ reconstruct grid map from $F_{un}$ and $F_{occ}$
7: $c_k \in C_k = clustering(F_{un})$
8: $V_r = MLP(s_{ri})$    $V_c = MLP(c_k)$
9: $G_r, G_c, G_{rc} =$ create graph from $V_r, V_c$
10: **for** $l = 0$ *to layers*:
11:     attention weight update $V_r, V_c$ and $E_{rc}$
12: **end for**
13: get allocation $r_i \sim c_k \in C_k$ from $E_{rc}$
14: **for** $r_i \in R$ **do**:
15:     **if** $NUM(C_k) >$ threshold:
16:         cluster $p$ in $C_k$, get $f_{kj} \in C_k$
17:         get optimal route $\rho_k^*$ for the highest utility $U_k^*$
18:     **end if**
19:     navigation and control
20: **end for**
21: return control command to target point

If frontier points are directly used as candidate targets for allocation, the action space remains too large, and the utility of frontier point aggregation is ignored. Therefore, frontier points are clustered using Mean-Shift, and the center of each cluster is used to calculate the robot's target, improving the efficiency of target allocation. Mean-Shift iteratively moves the kernel to a higher density region until it converges. However, the cluster center can easily be generated in an unreachable region, so the coordinates of the frontier points closest to the cluster center are selected as the candidate target point $c_k \in C_k \subset C$, where $c_k$ represents the center of cluster $C_k$, all the cluster of G formed the Fig. 2 illustrates the preprocessing pipeline.

*B. Target Allocation With mGNN*

Through the reconstructed grid occupancy map, the cluster information and the location information of the robot, the target point assignment problem can be described as the correlation problem between the robot node and the target node. Inspired by [17], this correlation is represented by a bipartite graph. Specifically, a cross graph $G_{rc}$ is established between the node $V_r$ of the robot graph $G_r$ and the node $V_c$ of the cluster center graph $G_c$, and the utility edge $E_{rc}$ is estimated by the multi-graph neural network based on the attention mechanism. Fig. 4 gives the pipeline of the global planning.

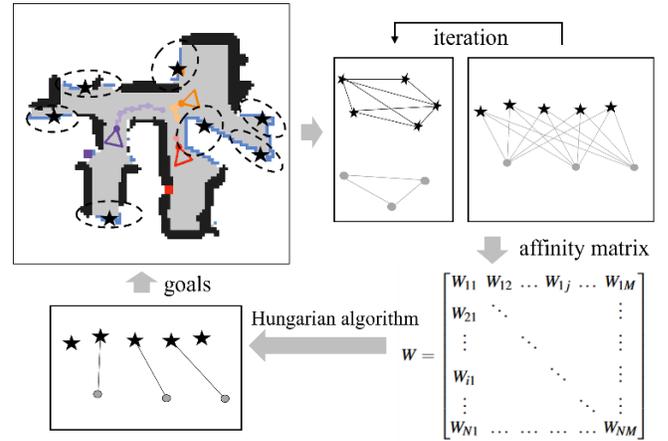

Fig. 4.     Global Planner With Robots And Clusters

Specifically, for $s_{ri} \in R^3$ and $c_k \in R^3$, a Multi-Layer Perception (MLP) is built to extract the high-dimensional features $v_{r\_i}^{(l)} \in R^{32}$. And calculate the query $q_{r\_i}^{(l)} \in R^{32}$, key $k_{r\_i}^{(l)} \in R^{32}$ and value $u_{r\_i}^{(l)} \in R^{32}$ of the node by three linear projections $L_{query}, L_{key}, L_{value}$, and let the similarity $sim_{r\_ij}^{(l)} = \left( q_{r\_i}^{(l)} \cdot k_{r\_j}^{(l)} \right)$. From this we can calculate the attention weight from node $j$ to node $i$:

$$e_{r\_ij}^{(l)} = \frac{\exp\left(sim_{r\_ij}^{(l)}\right)}{\sum_{h:(i,h)\in E_r} \exp\left(sim_{r\_ih}^{(l)}\right)}, \quad (4)$$

then the attention value $A_{r\_i}^{(l)}$ can be calculated by:

$$A_{r\_i}^{(l)} = \sum_{h:(i,h)\in E_r} e_{r\_ih}^{(l)} u_{r\_h}^{(l)}, \quad (5)$$

accordingly, the node features can be updated by MLP $f_V$:

$$v_{r\_i}^{(l+1)} = v_{r\_i}^{(l)} + f_V\left(v_{r\_i}^{(l)}, A_{r\_i}^{(l)}\right), \quad (6)$$

the same process is performed for the candidate target point graph $G_c$ to update the node features. After completing the node feature update of the self-graph, the feature update operation of the cross-graph can be performed. The shortest distance $d_{ik}$ between nodes $r_i$ and $c_k$ is obtained by using the $A^*$ algorithm in the occupancy grid map, the information gain from $i$ to $k$ $g_{ik}$ is represented by the number of members of the corresponding cluster $|C_k|$. These two pieces of information are considered in the calculation of the directed edge features from node $i$ to $k$. In cross graph operation, the similarity is represented as a multi-layer perception $sim_{rc\_ik}^{(l)} = f_{edge}\left(q_{r\_i}^{(l)}, k_{r\_k}^{(l)}, d_{ik}, g_{ik}\right)$, and the edge attention weight is calculated by:

$$e_{rc\_ik}^{(l)} = \frac{\exp\left(sim_{rc\_ik}^{(l)}\right)}{\sum_{h:(i,h)\in E_r} \exp\left(sim_{rc\_ih}^{(l)}\right)}, \quad (7)$$

and the attention value $A_{rc\_ik}^{(l)}$ can be calculated by

$$A_{rc\_ik}^{(l)} = \sum_{h:(i,h)\in E_{rc}} e_{rc\_ik}^{(l)} u_{c\_h}^{(l)}, \quad (8)$$

then the node feature will be secondary updated by replacing the attention value in (6) with $A_{rc\_ik}^{(l)}$. The self-graph and cross-graph sequentially constitute an operation block, and each time l layers of operation blocks are used to calculate the features of robot nodes and cluster center nodes.

Our graph operations are executed at the computing center, utilizing Proximal Policy Optimization (PPO) as the policy optimizer. Specifically, the reward value will be expressed in terms of the consumption time step $R_{time}$ and the exploration area increase $R_{area}$, where the former is a fixed negative real number and the latter is $R_{area} = A(G) - A(G_{-1})$, where the A is the area of occupancy grid map at time t, therefore the entire reward is:

$$R = R_{time} + \lambda R_{area}, \quad (9)$$

where the $\lambda$ is the hyper-parameter to balance the two rewards. After the graph operation is completed, a weight matrix is obtained, which is the preference degree of robot $r_i$ to go to cluster center $c_k$. this point, the problem becomes one of allocating cluster center nodes to each robot node in the bipartite graph $G_{rc}$ to maximize allocation utility. In this paper, the Hungarian algorithm [20] is used for calculating the $r_i \sim c_k$. Since the Hungarian algorithm requires the same number of frontier points as target points, virtual robots with negative infinity weights are added and subsequently skipped in the final output.

*C. Utility-Based Local Planning*

After obtaining the cluster centers, merely visiting these centers does not suffice for local exploration in large scenes. Inspired by [18], local routing planning and control planning are carried out when the average number of cluster members reaches the threshold $\theta$.

For robot $r_i$ and its assigned cluster center $C_k$, its members are clustered by Mean-Shift algorithm to get local candidate target points $f_{kj} \in C_k$. To compute local exploration routes efficiently, both gain and cost should be considered, therefore we propose a routing scheme that identifies the optimal route to visit all local candidate target points.

Define $\rho_k = [f_{k0}, f_{k1}, f_{k2}...f_{kn}]$ is the path of the local candidate point, where $f_{k0}$ is the current position of robot $r_i$, $G_{kj}$ is the information gain of $f_{kj}$, which is the number of members of the subclass, $L_{kj}$ is the distance from $f_{k0}$ to $f_{kj}$, calculated by the $A^*$ algorithm [21], and the purpose is to obtain an optimal path $\rho_k^* = [f_{k0}, f_k^*]$. This maximizes the following overall utility function:

$$U_{ik}^* = MAX \sum_{j=1}^{n} (G_{kj} + c \cdot L_{kj}), \quad (10)$$

where c is the balancing factor used to balance the two benefits. This problem is similar to the Travelling Salesman Problem (TSP) problem, but the standard TSP only considers the distance cost without considering the information gain, and it needs to return to the starting point after all the goal point exploration is completed. Many approaches typically use the number of unknown areas or frontiers around the target point as a measure of information gain. While this can provide substantial benefits, it often leads to a preference for targets with higher information gain because the sensor's scanning range can influence nearby candidate targets. Therefore, the following function is calculated prior to determining the route:

$$\begin{cases} G_{kj} = \exp\left(k_g \cdot \left(\frac{G_{kj} - \min(G_{kh})}{\max(G_{kh}) - \min(G_{kh}) + epsilon} - 1\right)\right) \cdot U_G \\ L_{kj} = \ln\left(k_l \cdot \left(\frac{L_{kj} - \min(L_{kh})}{\max(L_{kh}) - \min(L_{kh}) + epsilon} + 1\right)\right) \cdot U_L \end{cases}, \quad (11)$$

where $k_g$, $k_l$ are the rate of decay, $U_G$ and $U_L$ are the upper limit of normalization, and *epsilon* is a particularly small positive number to prevent division by zero. The purpose of this calculation is to enlarge the information gain gap and narrow the distance gap.

After preprocessing, subsequences need to be found for the best route. However, finding the subsequence can be challenging when there are many target points. Therefore, the method of reversing subsequence is utilized, specifically, (1) the descending order of $G_{kj}$ is used as the initial sorting, (2) randomly pick two non-robot indexes $x, y$, the subsequence between them is reversed to generate a new sequence, all possible cases are calculated, and the route with the highest utility is taken as the new initial sequence. (3) Repeat steps (1) and (2) until none of the reversed subsequences of some initial sequence yields a path with higher utility.

After calculating $\rho^*$, the navigation stack of the Robot Operating System (ROS) is used to navigate to the target point.

V. PREFORMANCE EVALUATION

*A. Experiment Setup*

**Experiment Environment:** We train on the iGibson physical simulator, use three scenes in the Gibson dataset for training and validation, and take three different scenarios of small, medium and large for verification. The Turtlebot robot is used as the agent, which is equipped with velodyne16 LiDAR, driven by a differential controller. During initialization, it is ensured that there is enough overlap area between robots to ensure accurate positioning between robots, and the default relative pose is known during this period, so the C-SLAM algorithm is not applicable for positioning. The ROS Navigation Stack is used for path planning and navigation.

TABEL I
THE NUMERICAL RESULTS
the percentage in brackets refers to the superiority of our method over the best method

| test | small | | middle | | large | |
| --- | --- | --- | --- | --- | --- | --- |
| method | Cov. | steps | Cov. | steps | Cov. | steps |
| Coscan | 99.7 | 874.8 | 99.9 | 924.8 | 99.4 | 1175.0 |
| NeuralComapping | 99.8 | 425.0 | 99.5 | 934.8 | 99.8 | 1524.8 |
| Ours | 98.9 | 369.8 (12.9% ↑) | 99.6 | 740.6 (19.9% ↑) | 99.8 | 1156.2 (1.6% ↑) |

**Parameters Setting:** For the training progress, every scene will be trained with 100 episodes, every episode consists of 25 steps. The resolution of the occupancy grid map is 0.05, which means each grid has a side length of 5 cm, and the scan range of the LiDAR is 6 m, the line velocity is limited in 2 m/s, the angel velocity is limited in 1.5 r/s; $\lambda = 2, c = k_g = k_l = 1, U_G = U_L = 5$. The clustering bandwidth threshold in the Global Planning Layer is automatically adjusted according to the acquired data and is set to the 15th percentile of the distance between the input samples as the bandwidth.

**Termination criterion:** The purpose of the exploration task is to completely scan the area, but in practical applications, the real range is not always known. And although most of the noise is removed in the early stage and preprocessing stage, some unreachable areas may still occur due to the uncertainty of the environment, so the termination condition is set as when there is no accessible candidate target. In addition, some small frontier points may be difficult to scan due to physical limitations, but the environment has almost been explored at this point, so the experiment is terminated when the number of all cluster members is less than a certain threshold (set to 2 in this experiment).

**Evaluation metrics:** Metrics for robotic exploration tasks include coverage of the environment as well as the time to complete exploration. Because time is limited by the environment, software and hardware, time steps are used as the evaluation index witch is represented by ROS time in robot operation system. The coverage (Cov.) is then the ratio of the explored area (the sum of free and occupied space) to the ground truth.

*B. Result*

We use three scenes from the Gibson dataset for training and validation. The validation scenes are categorized as small, medium, and large. Table 1 presents the results of our algorithm on the Gibson dataset for multi-robot autonomous exploration. Each scene was explored 10 times, and the result is reported as the mean value.

As shown in Table I, our method has been evaluated across three different scenarios after multiple trials. The table indicates that each method can approximately complete the construction of the map, and our method demonstrates significant advantages. By utilizing neural networks for long-term planning, considering the impact of clustering, and employing more rational routing strategies in local planning, our approach effectively avoids repeated exploration and invalid access. Consequently, compared to the baseline method, our method requires fewer steps to complete the exploration of the environment while achieving similar or better coverage. The enhanced collaboration is particularly beneficial in complex environments where efficient coordination among multiple robots is crucial for successful exploration and mapping.

Additionally, due to the sparsity of the data format we designed for communication, the data transmission volume between robots and the computing center is significantly reduced. As shown in Table II and Fig. 5, in our experimental scenarios, data transmission volume was reduced by more than 30%, demonstrating the effectiveness of our design. This reduction in data transmission not only improves the efficiency of the system but also reduces the load on the communication network, making our approach more scalable and robust in larger and more complex environments.

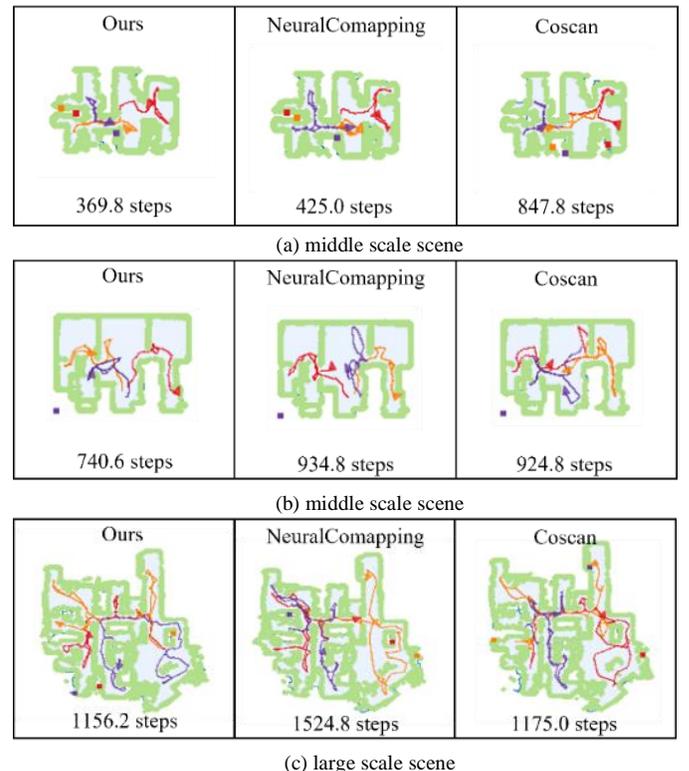

(a) middle scale scene

(b) middle scale scene

(c) large scale scene
Fig.5 visual results in each scene

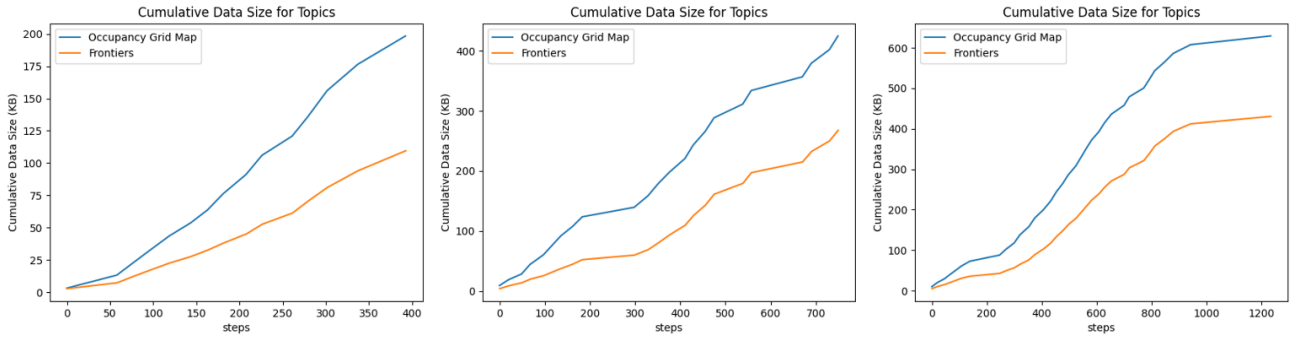

Fig.6 The amount of data transferred (traditional method based on grid map and ours based on frontiers)

TABEL II
THE REDUCED AMOUNT OF TRANSMITTED DATA

| scene \ type | Occupancy Grid Map | Frontiers | Reduction |
|---|---|---|---|
| Small | 201.6 KB | 102.5 KB | 49.16% ↓ |
| Middle | 417.4 KB | 265.8 KB | 36.32% ↓ |
| Large | 616.4 KB | 409.3 KB | 33.60% ↓ |

## VI. CONCLUSION

In this paper, to address the challenges in traditional robot exploration tasks, we propose a three-layer planning system architecture that combines learning-based and frontier-based methods. In the preprocessing layer, a sparse graph composed of frontier points with a closed internal area is obtained, enabling the computing center to reconstruct the grid map and reduce transmission load. The global planning layer achieves optimal allocation through a graph neural network by establishing undirected graph of robot nodes and cluster centers. The local planning layer provides a more reasonable route for local target points. The entire system is trained using deep reinforcement learning, and its superiority and generalization are validated through experiments. Although our approach can reduce the exploration time and the amount of data transmitted, it is still limited by the central architecture. In future work, multi-agent reinforcement learning and distributed schemes that more suitable for multi-robot systems could be considered to reduce dependence on the computing center.